\documentclass{article}
\pdfoutput=1
\usepackage{amsmath,graphicx,cipconf}

\toappear{4th International Workshop on Cognitive Information Processing, May 26--28, 2014, Copenhagen, Denmark}

\usepackage[hidelinks]{hyperref}

\usepackage{pgfplots}

\newcommand\copyrighttext{\footnotesize
The final version is published in \textit{Cognitive Information Processing (CIP), 2014 4th International Workshop on,} 1--6, 2014, \href{http://dx.doi.org/10.1109/CIP.2014.6844513}{doi: 10.1109/CIP.2014.6844513}.\\
\copyright\ 2014 IEEE. Personal use of this material is permitted. Permission from IEEE must be obtained for all other uses, in any current or future media, including reprinting/republishing this material for advertising or promotional purposes, creating new collective works, for resale or redistribution to servers or lists, or reuse of any copyrighted component of this work in other works.
}

\renewcommand\copyrightnotice{%
\begin{tikzpicture}[remember picture,overlay]
\node[anchor=south,yshift=10pt] at (current page.south) {{\parbox{\dimexpr\textwidth-\fboxsep-\fboxrule\relax}{\copyrighttext}}};
\end{tikzpicture}%
}

\usepackage{amsmath}
\usepackage{graphicx}
\usepackage{epstopdf}
\usepackage{subfigure}
\usepackage{gensymb}
\usepackage{url}
\usepackage{soul}
\usepackage{color}
\usepackage{framed}
\usepackage{flushend}

\newcommand{\at}{\makeatletter @\makeatother}

\title{Decoding index finger position from EEG using random forests}
\name{Sebastian Weichwald\textsuperscript{1}, Timm Meyer\textsuperscript{1}, Bernhard Sch\"olkopf\textsuperscript{1}, Tonio Ball\textsuperscript{2,3,$\ddagger$,$\dagger$}, Moritz Grosse-Wentrup\textsuperscript{1,$\ddagger$}\thanks{$\ddagger$ TB and MGW contributed equally to this study.}\thanks{$\dagger$ This work was partially funded by the DFG grant EXC 1086 BrainLinks-BrainTools to the University of Freiburg, Germany.}}
\address{
	\textsuperscript{1} Max Planck Institute for Intelligent Systems, T\"ubingen, Germany\\
	{\small\texttt{\{sweichwald,tmeyer,bs,moritzgw\}\at tuebingen.mpg.de}}\\
	\textsuperscript{2} Epilepsy Center, University Medical Center Freiburg, Freiburg, Germany\\
	{\small\texttt{tonio.ball\at uniklinik-freiburg.de}}\\
	\textsuperscript{3} Bernstein Center Freiburg, University of Freiburg, Freiburg, Germany
}
\begin{document}

\maketitle
\copyrightnotice

\begin{abstract}
While invasively recorded brain activity is known to provide detailed information on motor commands, it is an open question at what level of detail information about positions of body parts can be decoded from non-invasively acquired signals. In this work it is shown that index finger positions can be differentiated from non-invasive electroencephalographic (EEG) recordings in healthy human subjects. Using a leave-one-subject-out cross-validation procedure, a random forest distinguished different index finger positions on a numerical keyboard above chance-level accuracy. Among the different spectral features investigated, high $\beta$-power (20--30 Hz) over contralateral sensorimotor cortex carried most information about finger position. Thus, these findings indicate that finger position is in principle decodable from non-invasive features of brain activity that generalize across individuals.

\end{abstract}


\section{INTRODUCTION}
Invasive recordings of brain activity in human subjects have been shown to provide rich information on various aspects of motor control. For instance, single-unit recordings in primary motor cortex enable the control of end-effectors in multiple dimensions \cite{hochberg2006neuronal,hochberg2012reach}. Subdural electrocorticography (ECoG) allows for the decoding of arm movement direction \cite{leuthardt2004brain} as well as the laterality of index finger movement \cite{ball2004towards}, and can distinguish between movements of individual fingers of the same hand \cite{zanos2008electrocorticographic,miller2009decoupling}. It has even been shown that ECoG recordings provide information on the time course of the flexion of individual fingers \cite{acharya2010electrocorticographic,kubanek2009decoding}.
It is at present unclear, however, to which extent such information can also be decoded from non-invasive measures of brain activity. To date, empirical evidence has been presented that electroencephalography (EEG) and magnetoencephalography (MEG) recordings provide information on the laterality of index finger movements \cite{li2004classification,lehtonen2008online,kauhanen2006classification} as well as on arm movement trajectories \cite{georgopoulos2005magnetoencephalographic,jerbi2007coherent,waldert2008hand}. Furthermore, predictive features in MEG as well as EEG recordings for discriminating movements of individual fingers on one hand have been reported \cite{quandt2012single}.

Here, this line of work is extended by investigating whether index finger position can be decoded on a single-trial basis from EEG recordings. This is a particularly challenging task, as there exists an overlap of neurons in primary motor cortex controlling individual finger movements, and the encoding of muscle activations and joint positions does not follow a strict somatotopic organization \cite{schieber2001constraints}. This raises the question whether EEG recordings provide sufficient information to differentiate between different index finger positions.

Using a random forest classifier empirical evidence is presented that different index finger positions can indeed be differentiated on a single-trial basis from EEG recordings. The random forest classifier was trained with a leave-one-subject-out cross-validation procedure involving 20 healthy subjects,
indicating that the representation of index finger position is shared across subjects. The classification was primarily based on changes in high $\beta$-band power (20--30 Hz) over the contralateral motor cortex.

The remainder of this paper is structured as follows: In section II the experimental setup and paradigm, the recorded data, the computation of features, the classification procedure and the feature importance scores are described. Section III presents the results of the applied procedure while section IV closes with a discussion of the results and points to some interesting questions for further investigation.
\section{METHODS}
\subsection{Experimental Paradigm}

\begin{figure}[t]
    \centering
    \includegraphics[bb=0 0 441 519,keepaspectratio=true,width=.4\textwidth]{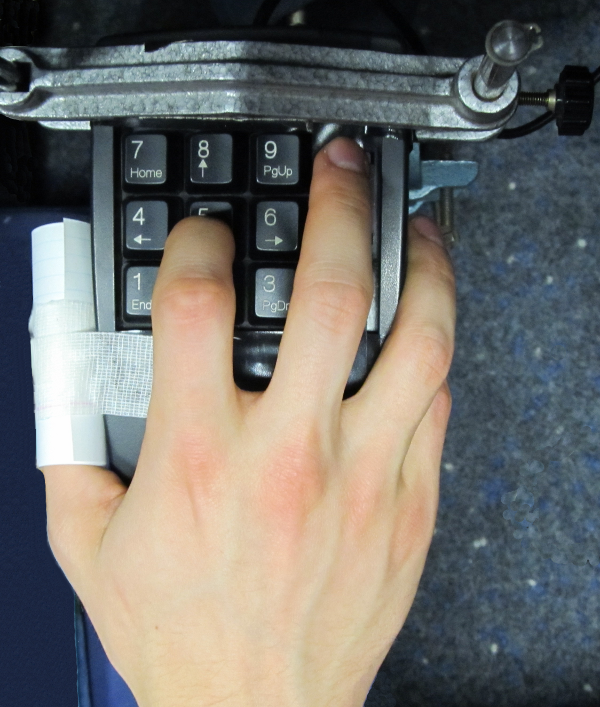}
    \caption{Experimental setup: The index finger was used to press keys on a 3x3-keyboard attached to an arm rest. Thumb position is fixed by a tube. Visual cover is removed here.}
    \label{fig:setup}
\end{figure}

During the experiment subjects were seated in a comfortable chair approximately 1.25 m in front of a computer screen. A numeric 3x3-keyboard was attached to the chair's right arm rest (cf.~fig.~\ref{fig:setup}). Subjects placed their right arm comfortably on the arm rest, so that they were able to easily reach all nine keys of the keyboard with their index finger and without moving their arm or wrist. To keep the hand position fixed throughout the experiment, subjects were asked to keep their thumb in a tube attached to the side of the keyboard. The keyboard was blocked from the subjects' view by a black wooden board.

The experimental paradigm consisted of three different states: a five minute resting state, during which the subjects were asked to fixate a gray cross displayed centrally on the screen; a trial state of approximately five minutes, during which subjects were familiarized with the experimental paradigm; and the actual testing state, which consisted of 15~sessions of 90~trials each.

\begin{figure}[t]
    \hspace*{\fill}
    \subfigure[]{
        \label{fig:stimuli-r}
        \includegraphics[natwidth=122,natheight=122,height=20mm,keepaspectratio=true]{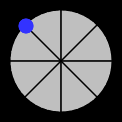}
    }
    \hfill
    \subfigure[]{
        \label{fig:stimuli-w}
        \includegraphics[natwidth=122,natheight=122,height=20mm,keepaspectratio=true]{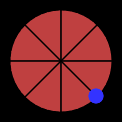}
    }
    \hfill
    \subfigure[]{
        \label{fig:stimuli-map}
        \includegraphics[natwidth=122,natheight=122,height=20mm,keepaspectratio=true]{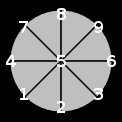}
    }
    \hspace*{\fill}
    \caption{Visual stimuli: \ref{fig:stimuli-r} indicates the current target key; \ref{fig:stimuli-w} indicates that the currently pressed key is incorrect. Figure~\ref{fig:stimuli-map} shows how the class labels/key positions are mapped to the different visual stimuli.}
    \label{fig:stimuli}
\end{figure}

All visual stimuli (cf.~fig.~\ref{fig:stimuli}) were presented within a visual angle of less than 2\degree. The target key was indicated by the position of a blue dot superimposed on a grey circle (cf.~fig.~\ref{fig:stimuli-r}). Figure~\ref{fig:stimuli-map} shows the mapping of class labels and key positions to the different visual stimuli. The grey circle served as a fixation cross, and lit up in red if the wrong key was pressed (cf.~fig.~\ref{fig:stimuli-w}).

Within each trial the target key needed to be pressed for a duration between three and four seconds, randomly drawn from a uniform distribution. If the subject released the target key too early, the timer was restarted. If the target key was pressed sufficiently long, the next target key was presented, prompting the subject to release the current- and press the new target key.

Each of the 15 sessions consisted of 90 randomly permuted trials, avoiding repetitions of the same target key and ensuring ten trials per target key within each session. This resulted in 150 trials for each target key per subject.

\subsection{Subjects}
Twenty healthy subjects (mean age of 28.4 years with a standard deviation of 4.1 years) participated in this study, all of which indicated that they were right-handed (laterality scores according to the Edinburgh inventory \cite{oldfield1971assessment} between~$\pm$0 and~+100, mean +68, standard deviation 26). Each subject gave informed consent in agreement with guidelines set by the Max Planck Society.

\subsection{Experimental Data}
During the experiment EEG was recorded at 121 active electrodes, placed according to the extended international 10-20 system with a sampling rate of 2 kHz using a QuickAmp amplifier (Brain Products GmbH, Gilching, Germany) with built-in common average reference. Impedances were ensured to be below 25 k$\Omega$ at the beginning of each recording. Channels exceeding this threshold were switched off by manually connecting them to the ground channel of the amplifier.

To prepare the data for classification the following procedure was employed. First, the data of each subject was spatially filtered using a Laplacian setup \cite{mcfarland1997spatial}. Then, channels that were not present in every individual recording were removed, such that 106 channels common to all subjects remained. The data was then highpass filtered at a cut-off frequency of 3 Hz using a phase-preserving third-order Butterworth filter, and a common average reference filter was applied.

Only the first three seconds, i.~e.~6k samples, of each trial were used for further analysis. Due to quitting the experiment early, there is one trial missing for one subject and 300 trials missing for another subject. Thus a total of 26819 trials were recorded.

\subsection{Feature Computation}

For each subject, the $\mu$-band was identified manually by selecting the frequency band (width across subjects between 1~Hz and 3 Hz) with the highest bandpower in the 8--15 Hz range of channels over the left and right sensorimotor cortex using the resting state data only.

Then for each trial, the log-bandpower of the individual \mbox{$\mu$-band} and the log-bandpower of a general $\beta$-band \mbox{(20--30~Hz)} was computed for the whole trial as well as for 41 sliding windows of one second length, with the center of the window shifted in steps of 50ms, using a FFT with a Hann window. This yielded 84 bandpower features per trial and channel, resulting in 8904 features per trial in total.

For each subject feature values exceeding three standard deviations were marked as outliers. Specifically, the following step was iterated until no further outliers were found: Mark feature values which differ more than three standard deviations from the mean as outliers, where standard deviation and mean are calculated ignoring already marked outliers. Features marked as outliers were set to non-informational values in the training procedure (c.\,f.~section~\ref{sec:classif}).

Finally, the features not marked as outliers of each subject were normalised by subtracting the mean and dividing by the standard deviation over trials.

\subsection{Classification Procedure}\label{sec:classif}
A random forest classifier (RF) is used for decoding in this work \cite{breiman2001random}.
A RF combines multiple decision trees in an ensemble. The predicition is obtained by a voting scheme, i.\,e. a RF predicts the class most trees vote for. 
Every tree is grown on a random subset of training samples.
At each tree node the best split, determined by the class separability, of a random selection of features is chosen.
Thus, correlation between trees in a RF is reduced using bagging and random selection of features.

Using the features described in the previous section the following leave-one-subject-out cross-validation was employed. In contrast to the commonly employed within-subject cross-validation, this approach searched for features that generalize across subjects. In each fold, only the trials of 19 subjects were used to train a RF model \cite{breiman2001random} while the trials of the remaining subject were used as a test set. Values, that where previously marked as outliers, were set to non-informational values, i.~e.~to the mean of the particular feature in all training trials.

For each RF model 900 trees were grown using bagging, i.~e.~for every tree a training subset (63.2\% of the whole training set) was drawn by sampling with replacement; for each tree the remaining training data is called out-of-bag (OOB). While growing the trees, at each node the best split of $\lfloor \sqrt{8904} \rfloor = 94$ (as suggested in \cite{breiman2002manual}) randomly chosen features was used. For each RF model an OOB error estimate can be obtained by voting on the training data, while every tree is only allowed to vote on the corresponding OOB training data (cf.~\cite{breiman2001random}).

Each RF model was used to decode the target keys of the held-out subject and the prediction accuracy (PA) for each fold was calculated. In case of an ambiguous majority vote, a class with the maximum number of votes was assigned to this trial randomly.

The overall prediction accuracy of the cross-validation procedure was then estimated by dividing the total number of hits by the total number of predicted trials. A binomial test was employed to test the statistical significance of the cross-validated prediction result.

\subsection{Importance Scores}\label{sec:impscores}
For every RF model, feature importance scores were computed as described in \cite{breiman2001random}, i.~e.~the average percentual increase of the OOB error estimate when permuting one feature along trials is the importance score of that particular feature. Averaging these scores across all 20 RF models gives an overall feature importance score (FIS) for every feature.

To obtain an importance score for each channel (channel importance score, CIS) the FISs were mapped to the corresponding channels and added up. For each channel two scores, CIS$\mu$ and CIS$\beta$, were calculated by either only using the FISs of $\mu$-band features or $\beta$-band features, respectively.

To investigate dynamic changes in the importance of the $\mu$-band and $\beta$-band over the course of a trial, the FISs belonging to the channel with the highest channel importance score were used. This resulted in importance scores of the $\mu$-band (WIS$\mu$) and the $\beta$-band (WIS$\beta$) for the 41 time windows, and importance scores of the $\mu$-band (IS$\mu$) and the $\beta$-band (IS$\beta$) for the whole trial.

\section{RESULTS}
\begin{table*}[t]
    \small
    \centering
	\renewcommand{\arraystretch}{1.3}
	\caption{Prediction accuracy (PA) for each subject of the leave-one-subject-out cross-validation.}
	\label{table_example}
	\centering
	\tabcolsep=0.05cm
	\begin{tabular}{|c||c|c|c|c|c|c|c|c|c|c|c|c|c|c|c|c|c|c|c|c|}
		\hline
		Subject & 1 & 2 & 3 &  4 & 5 & 6 & 7 & 8 & 9 & 10 & 11 & 12 & 13 & 14 & 15 & 16 & 17 & 18 & 19 & 20 \\
		\hline
		PA in \% & 11.33 & 12.74 & 10.89 & 12.81 & 11.70 & 13.48 & 10.96 & 12.37 & 12.89 & 11.26 & 12.52 & 14.67 & ~9.91 & 11.78 & 12.01 & 13.41 & 12.89 & 13.48 & 11.70 & 12.59 \\
		\hline
	\end{tabular}
	\label{tab:pape}
\end{table*}

Classification accuracies achieved for each subject are shown in table~\ref{tab:pape}. In total, 3295 of 26819 trials were correctly predicted (12.29\%, $p < 8\cdot10^{-10}$, $N = 26819$).

\begin{figure}[t]
    \centering
    \includegraphics[keepaspectratio=true,width=.5\textwidth]{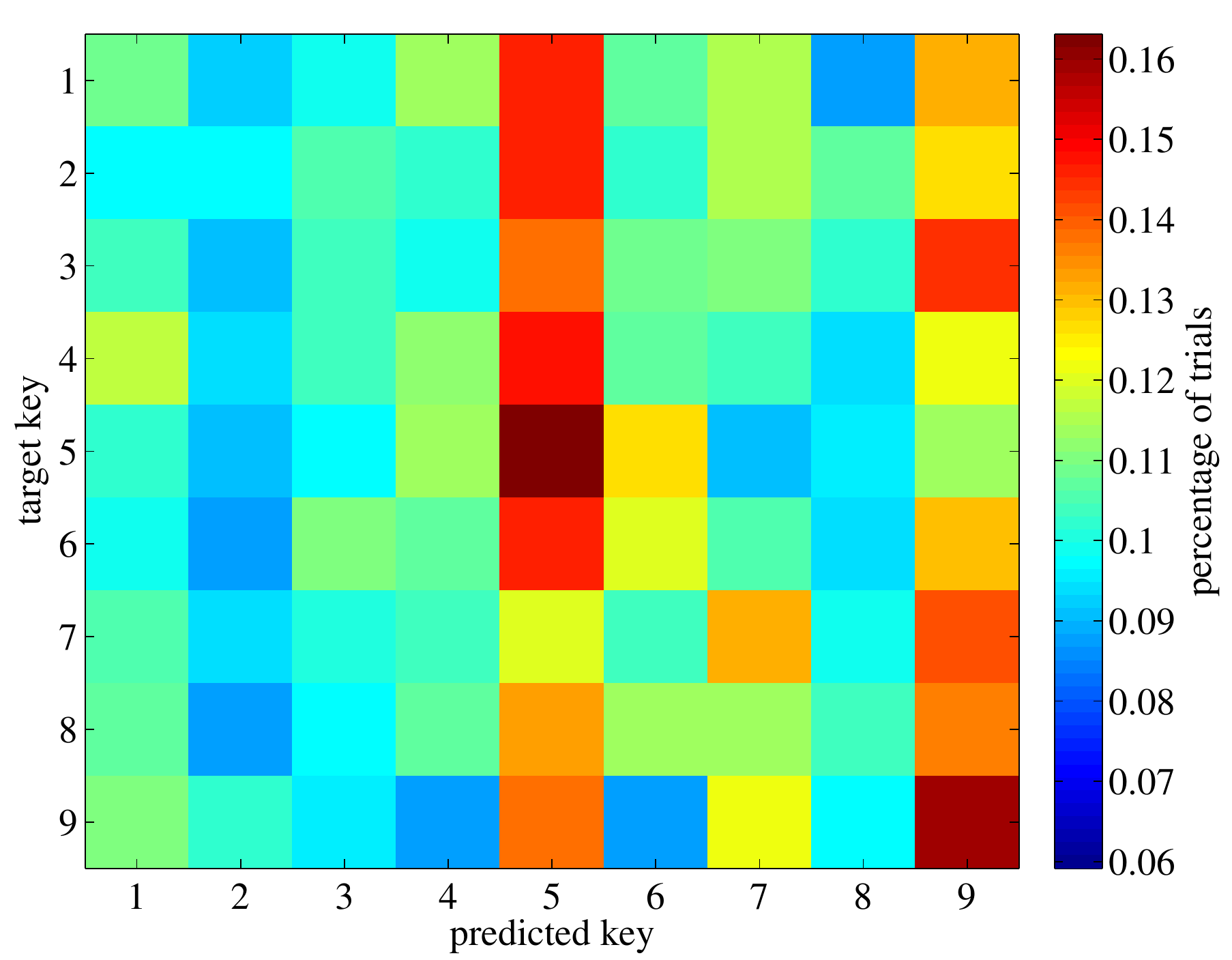}
    \caption{Confusion matrix showing the distribution of decoded positions for each actual target key. Values indicate percentage of trials. The colormap is adjusted so that green is indicating the chance level of 11.11\%.}
    \label{fig:confusion}
\end{figure}

\begin{figure}[t]
    \centering
    \includegraphics[keepaspectratio=true,width=.5\textwidth]{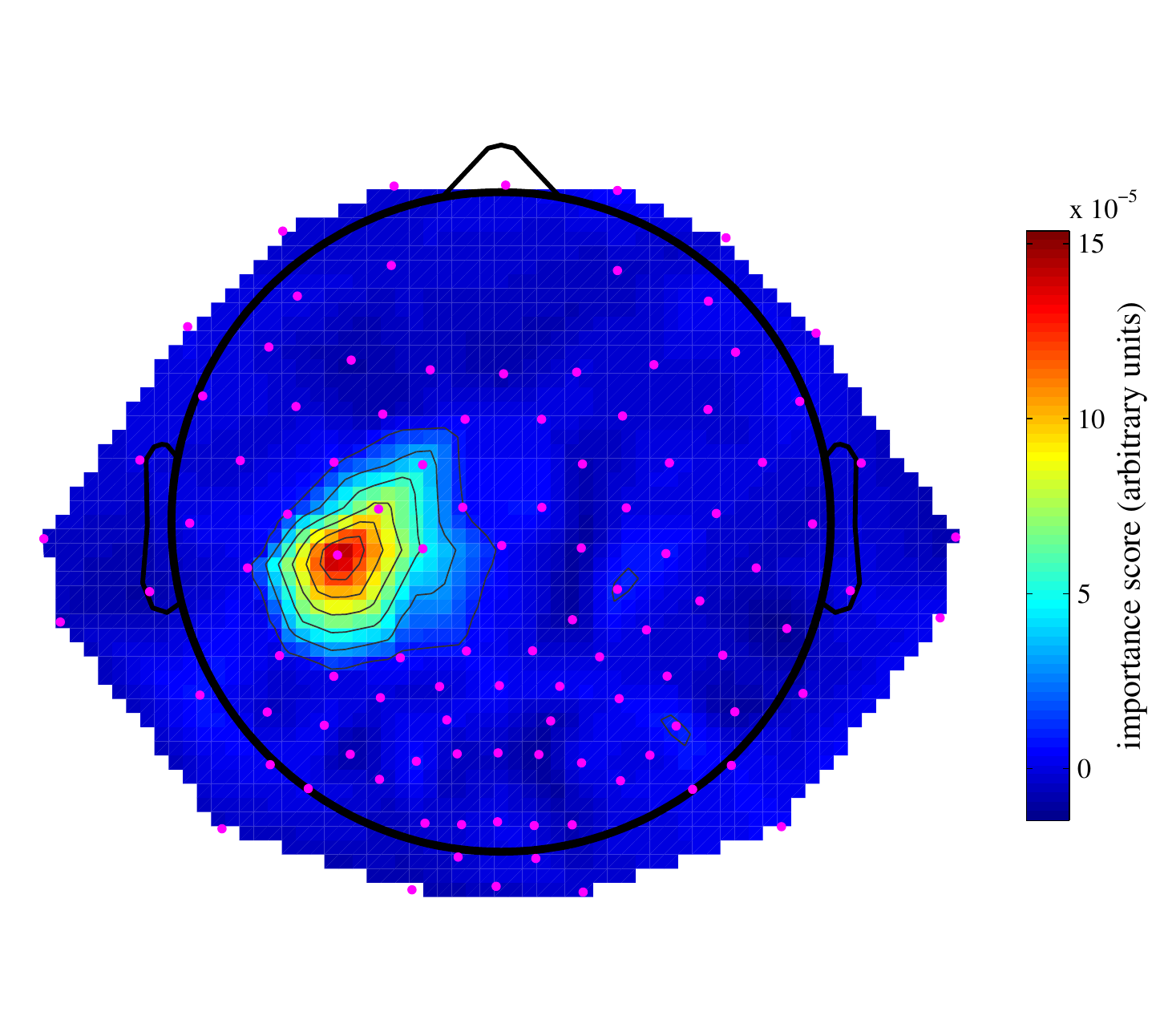}
    \caption{Topographic map of channel importance scores of the $\beta$-band (CIS$\beta$, cf.~\ref{sec:impscores}). Magenta dots denote EEG electrode positions. Maximum of the map is at C3.}
    \label{fig:impSum-beta}
\end{figure}

\begin{figure}[h!]
    \centering
    \includegraphics[keepaspectratio=true,width=.5\textwidth]{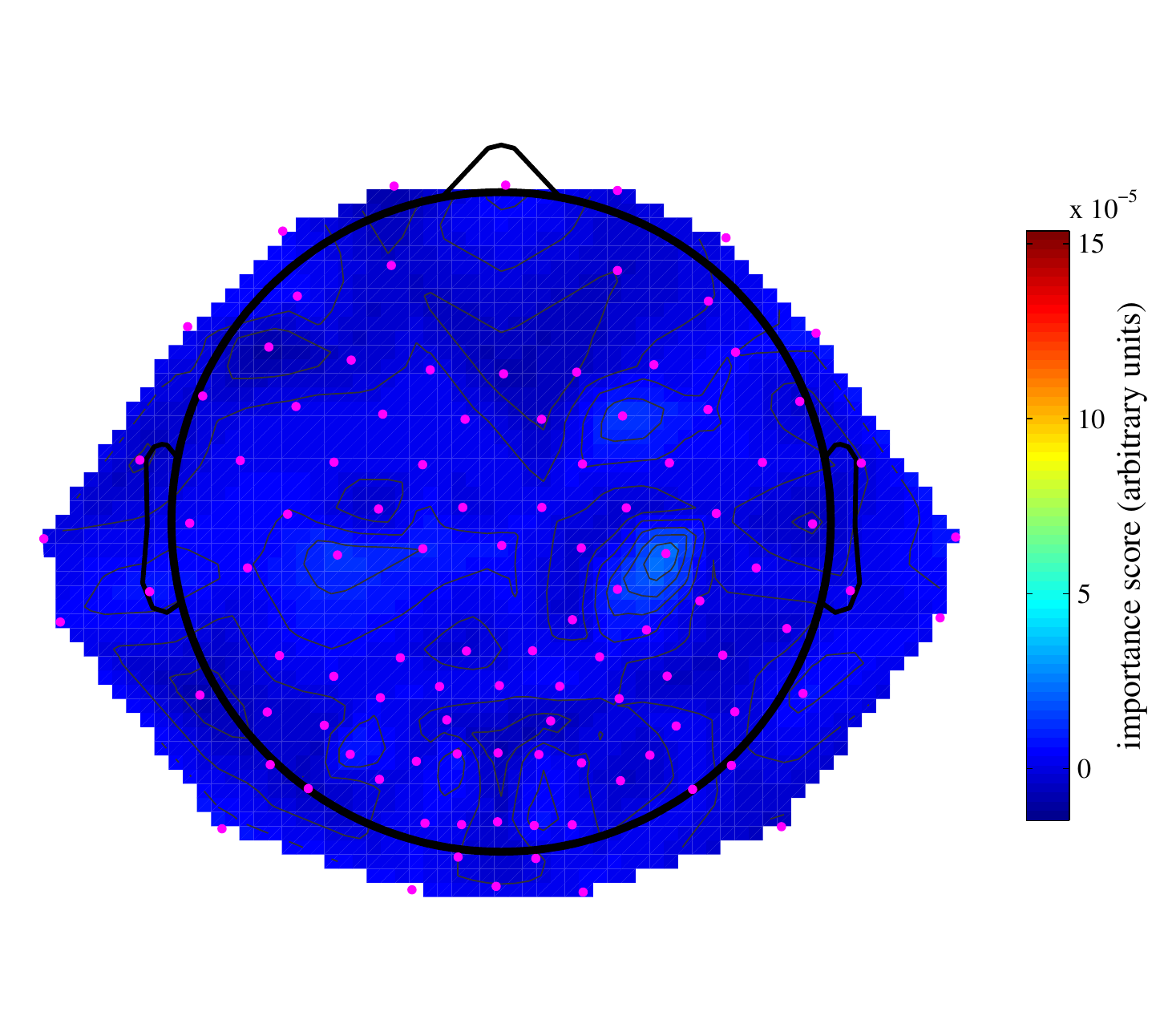}
    \caption{Topographic map of channel importance scores of the $\mu$-band (CIS$\mu$, cf.~\ref{sec:impscores}). Magenta dots denote EEG electrode positions.}
    \label{fig:impSum-mu}
\end{figure}

\begin{figure}[t]
    \centering
    \includegraphics[keepaspectratio=true,width=.5\textwidth]{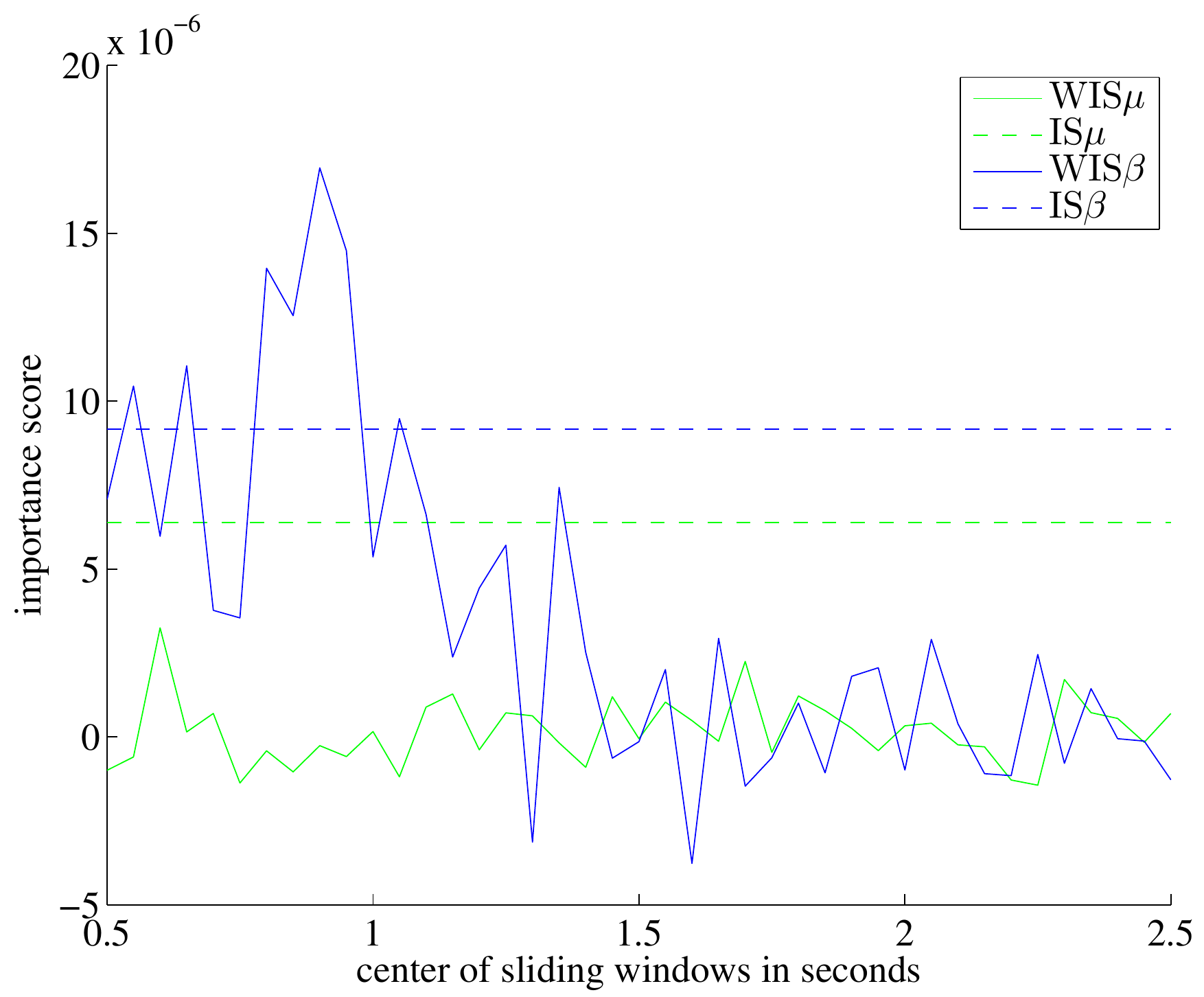}
    \caption{Importance scores of the 41 sliding windows obtained from the corresponding $\mu$-band feature (WIS$\mu$) and \mbox{$\beta$-band} feature (WIS$\beta$) importance scores; importance scores of the whole-trial $\mu$-band feature (IS$\mu$) and $\beta$-band feature (IS$\beta$) are shown as constant dashed lines; importance scores correspond to feature importance scores at C3.}
    \label{fig:slideImp}
\end{figure}

The confusion matrix in figure \ref{fig:confusion} shows which keys were predicted for each actual target key: target 5 (16.71\%) and target 9 (15.94\%) were most accurately predicted. For most target keys a preference for predicting the trial as a target 5 or target 9 trial can be observed. Table~\ref{tab:hits} shows for each key the true positive rate, i.\,e. correctly decoding trials with that target key, and the true negative rate, i.\,e. correctly rejecting trials with other target keys as not belonging to that class.

The topographic map of CIS$\beta$ (cf.~fig.~\ref{fig:impSum-beta}) identifies channels on the contralateral sensorimotor cortex as most predictive ones. CIS$\beta$ for C3 exceeds the mean of CIS$\beta$ by more than seven standard deviations and is higher than all CIS$\beta$ and CIS$\mu$. Therefore the importance scores WIS$\mu$, WIS$\beta$ and IS$\mu$, IS$\beta$ correspond to the FISs at C3 as described in section \ref{sec:impscores} and are shown in figure~\ref{fig:slideImp}.

The first windows centered at 0.5 to 1 (which are the windows from 0--1 seconds to 1--2 seconds) have up to 15 times higher WIS$\beta$ than the later windows. IS$\beta$ is higher than the mean of all WIS$\beta$ and higher than IS$\mu$.

In general the importance scores are smaller for the $\mu$-band than for the $\beta$-band; for comparison figures \ref{fig:impSum-beta} and \ref{fig:impSum-mu} show topographic maps of CIS$\beta$ and CIS$\mu$ on the same color scale. In addition to the contralateral channels, CIS$\mu$ shows predictive channels on the ipsilateral sensorimotor cortex (cf.~fig.~\ref{fig:impSum-mu}). CIS$\mu$ for C4 is more than four standard deviations higher than the mean. The whole trial score IS$\mu$ is much higher than all WIS$\mu$, which in contrast to WIS$\beta$ do not show any prominent peak (cf.~fig.~\ref{fig:slideImp}).

\begin{table}[t]
    \small
    \centering
	\renewcommand{\arraystretch}{1.3}
	\caption{True positive rate (tp) and true negative rate (tn) for each target key.}
	\label{table_example}
	\centering
	\tabcolsep=0.05cm
	\begin{tabular}{|c||c|c|c|c|c|c|c|c|c|}
		\hline
		Key & 1 & 2 & 3 &  4 & 5 & 6 & 7 & 8 & 9 \\
		\hline
		tp in \% & 10.94 & 9.80 & 10.40 & 11.14 & 16.71 & 12.01 & 13.22 & 10.40 & 15.94 \\
		\hline
		tn in \% & 89.48 & 90.77 & 89.89 & 89.58 & 86.06 & 89.28 & 89.07 & 90.27 & 86.92 \\
		\hline
	\end{tabular}
	\label{tab:hits}
\end{table}
\section{DISCUSSION}
In this paper, empirical evidence has been presented that discrete index finger positions are decodable from brain activity recorded by means of non-invasive EEG. $\beta$-band features strongly focalized over the contralateral sensorimotor cortex during the first second of each trial were found to be most predictive.

During the task employed in this study, a stable force needed to be maintained to press down the current target key and keep the finger in a fixed posture, which involves continuous tonic muscle activity. Therefore, it is unclear if the brain activity related to position-dependent differences in tonic muscle activity was decoded, or if in fact the current index finger position itself was decoded. As the WIS$\beta$ showed a strong preference for the first second of a trial, it can be argued that it is not a continuous brain activation pattern that is being decoded, but rather that a spatial updating \emph{after} a change in position, as manifested in a $\beta$-rebound \cite{pfurtscheller1996post}, provided information on the current target key. Thus, the decoding results described here on the level of finger movements may have a similar basis as previous findings on arm movements, as in both cases there was a prominent contribution from differences between a central resting position and the remaining
peripheral positions \cite{demandt2012reaching}. Further investigation is necessary to differentiate between these explanations, e.~g.~by training models on selected subsets of features. In addition, these findings raise the question whether the dynamics of $\beta$-activity or the $\beta$-rebound enable position decoding of other body parts as well.

Interestingly, significant classification accuracy was \sloppy achieved within a leave-one-subject-out setup using a random forest. This indicates that a pattern was successfully decoded that was consistent across subjects.
Furthermore, no parameter tuning, e.~g.~optimizing the number of trees or the number of variables to select the best split at each node, was carried out. It is likely that prediction accuracies can be further increased by tuning these parameters or training models on a subset of most important features.

An interesting feature of RF models, being ensemble methods, is their ability to handle large feature spaces. This enabled the extension of the feature space to multiple time windows, providing interesting insights into the dynamics of brain rhythms associated with index finger positioning. To date, non-linear classifiers have not been shown to outperform linear methods in brain-state decoding \cite{muller2003linear}. It remains to be seen whether the non-linearity of RF models contributed to the present decoding results, or if similar results can also be achieved by linear (ensemble) methods.

\bibliographystyle{IEEEbib}
\bibliography{bibfile}

\end{document}